\documentclass[10pt,twocolumn,letterpaper]{article}

\usepackage{iccv}
\usepackage{times}
\usepackage{epsfig}
\usepackage{graphicx}
\usepackage{subcaption}
\usepackage{amsmath}
\usepackage{amssymb}
\usepackage{multirow}

\iccvfinalcopy 

\def\iccvPaperID{14} 

\ificcvfinal\fi

\begin{document}

\title{Winning the ICCV 2019 Learning to Drive Challenge}

\author{Michael Diodato\textsuperscript{*}\\
Columbia University \\
{\tt\small mjd2249@columbia.edu}
\and
Yu Li\textsuperscript{*}\\
Columbia University \\
{\tt\small yl4019@columbia.edu}
\and
Manik Goyal\textsuperscript{1}\\
Columbia University \\
{\tt\small mg4106@columbia.edu}
\and
Iddo Drori\textsuperscript{2}\\
Columbia University \\
{\tt\small idrori@cs.columbia.edu}
}
\maketitle
\ificcvfinal\thispagestyle{empty}\fi

\makeatletter{\renewcommand*{\@makefnmark}{}
\footnotetext{* Columbia University Deep Learning Course Participants}\makeatother}

\makeatletter{\renewcommand*{\@makefnmark}{}
\footnotetext{1 Columbia University Deep Learning Course Assistant}\makeatother}

\makeatletter{\renewcommand*{\@makefnmark}{}
\footnotetext{2 Columbia University Deep Learning Course Instructor}\makeatother}

\begin{abstract}
Autonomous driving has a significant impact on society. Predicting vehicle trajectories, specifically, angle and speed, is important for safe and comfortable driving. This work focuses on fusing inputs from camera sensors and visual map data which lead to significant improvement in performance and plays a key role in winning the challenge. We use pre-trained CNN's for processing image frames, a neural network for fusing the image representation with visual map data, and train a sequence model for time series prediction. We demonstrate the best performing MSE angle and best performance overall, to win the ICCV 2019 Learning to Drive challenge. We make our models and code publicly available \cite{diodato2019learningtodrive}.
\end{abstract}

\section{Introduction}
\label{sec:introdcution}

Self driving cars have moved from driving in the desert terrain, spearheaded by the DARPA Grand Challenge, through highways, and into populated cities. End-to-end approaches directly maps input images to driving actions \cite{end2end}, predicting trajectories based on end-to-end deep learning models using supervised regression. Using an LSTM network, we predict both trajectory and speed from a set of input images and a semantic map. Recently, ChaeffeurNet \cite{bansal2018chauffeurnet} predicts trajectories and uses a mid-level controller to transfer these predictions to specific vehicles, avoiding the need to retrain a model for every different vehicle type. The use of visual semantic maps has shown to outperform the traditional end-to-end models which only used front-facing images as input data \cite{hecker2018end, hecker2019learning}. Recent work trains end-to-end models for predicting steering angle, speed, and driving trajectories. The use of LSTM models efficiently predicts steering \cite{fernando2017going} by taking into account long range dependencies. A single 3D CNN model which simultaneously performs detection, tracking and motion forecasting is also trained in an end-to-end fashion \cite{luo2018fast}. Since predicting steering angle alone is insufficient for self-driving, a multi-task learning framework \cite{yang2018end} is used to predict both speed and steering angle end-to-end. Augmenting existing data by using a pre-trained neural networks for image segmentation and optical flow estimation has shown to improve steering angle prediction \cite{hou2019learning}.

\begin{figure}
    \centering
        \includegraphics[width=0.495\linewidth]{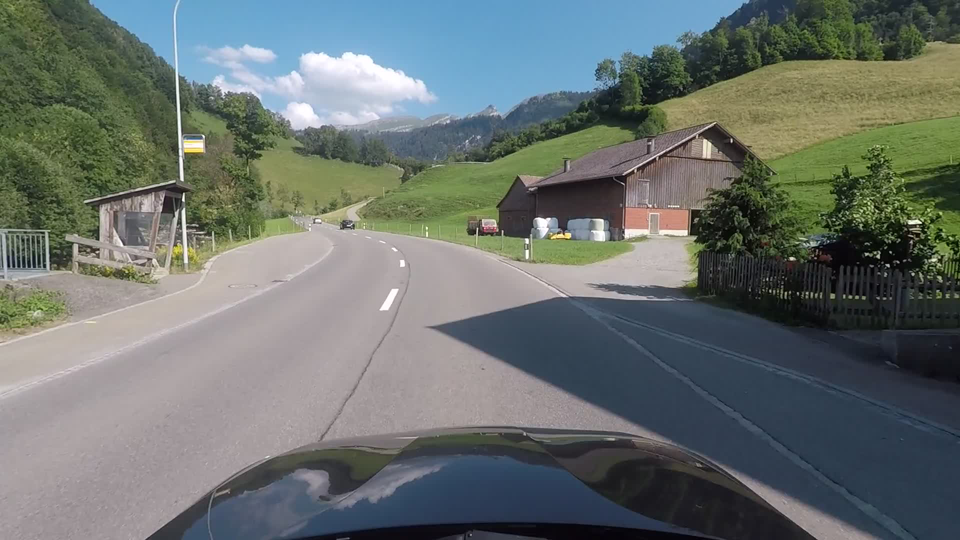}
        \includegraphics[width=0.495\linewidth]{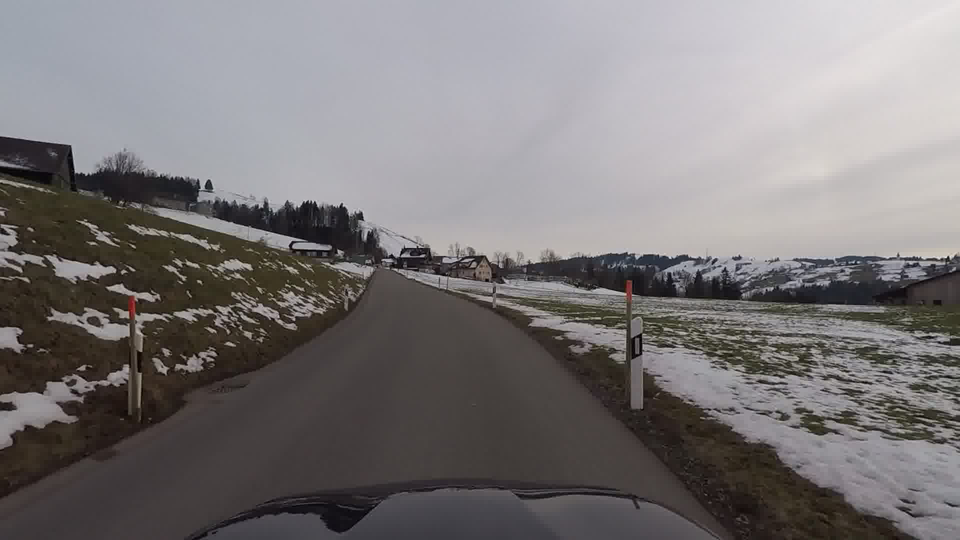}
        \includegraphics[width=0.495\linewidth]{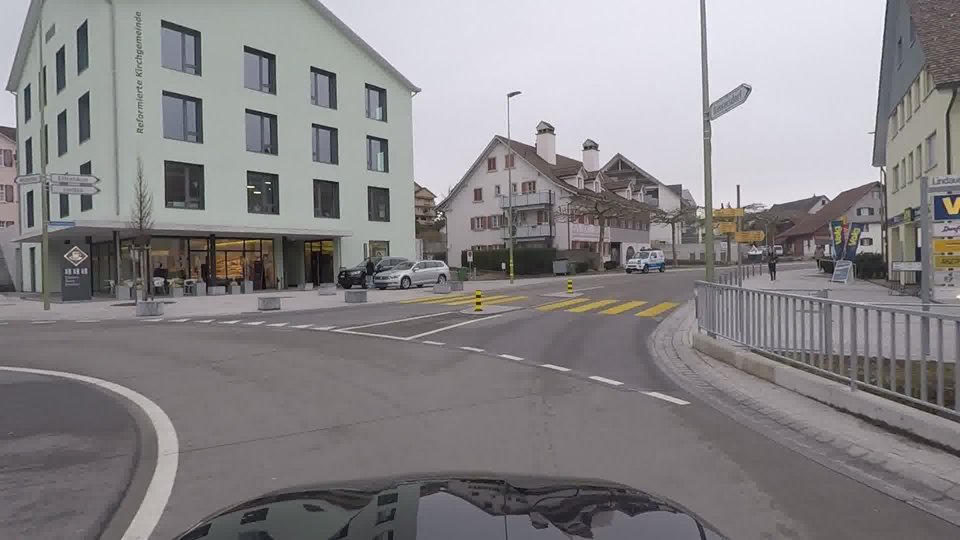}
        \includegraphics[width=0.495\linewidth]{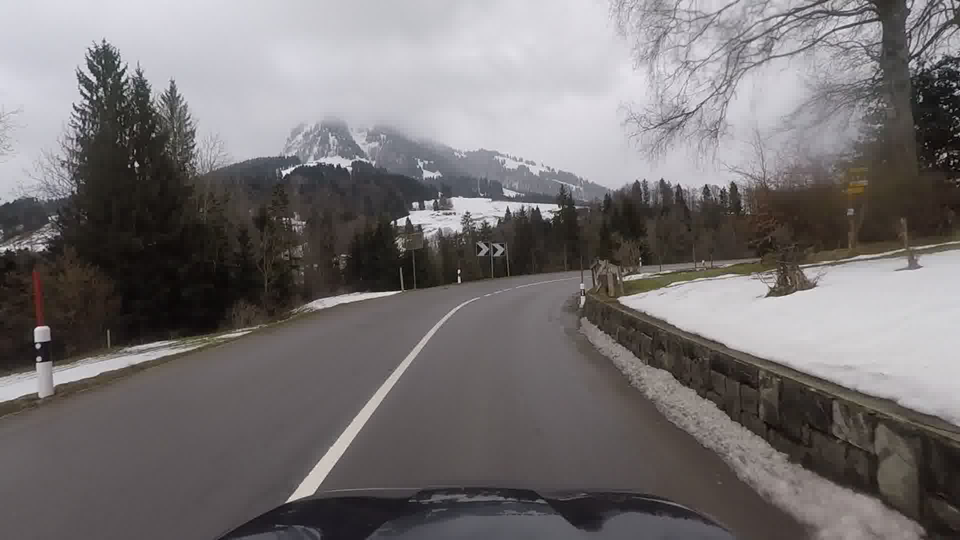}
        \includegraphics[width=0.495\linewidth]{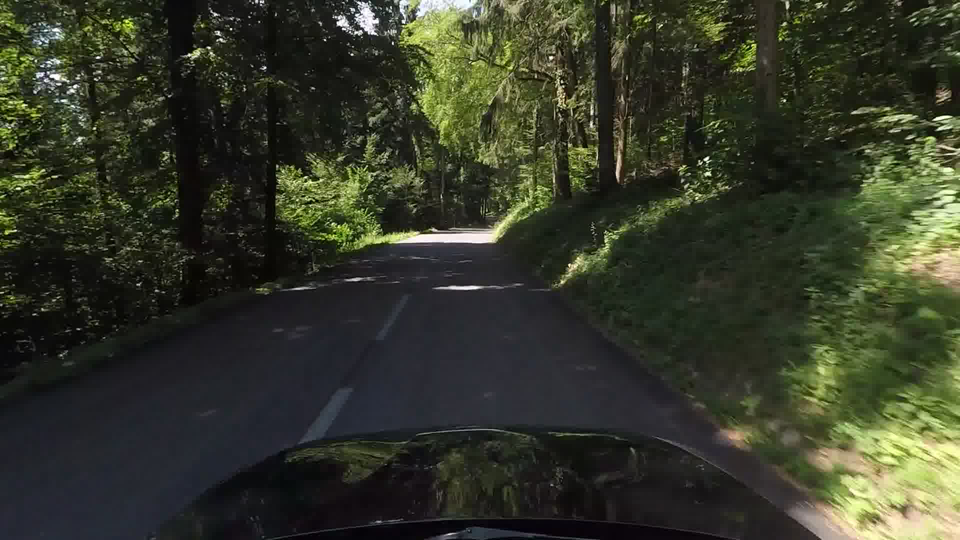}
        \includegraphics[width=0.495\linewidth]{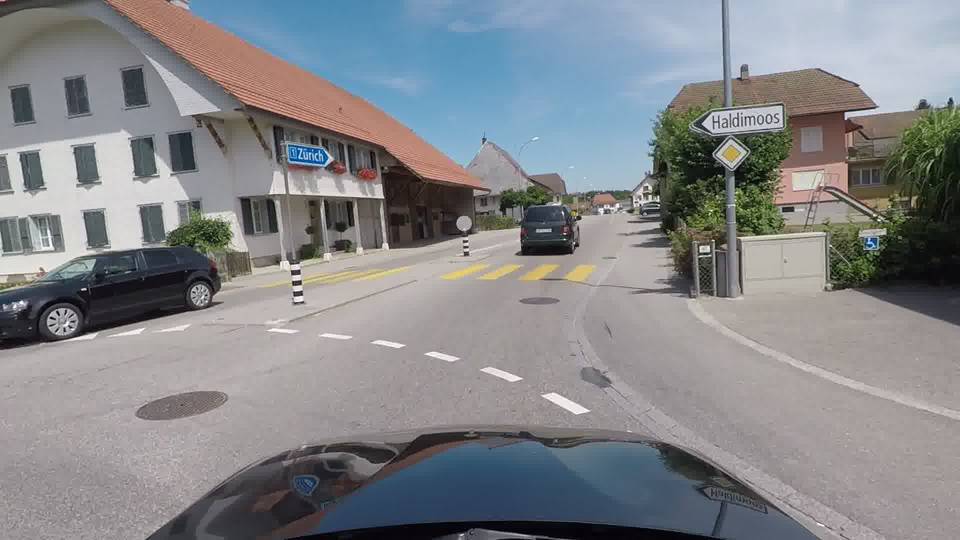}
        \includegraphics[width=0.495\linewidth]{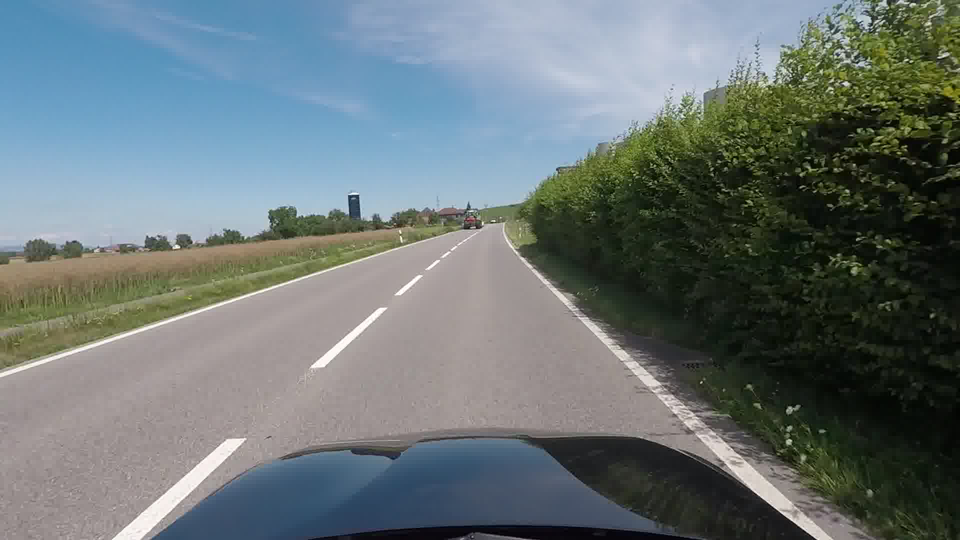}
        \includegraphics[width=0.495\linewidth]{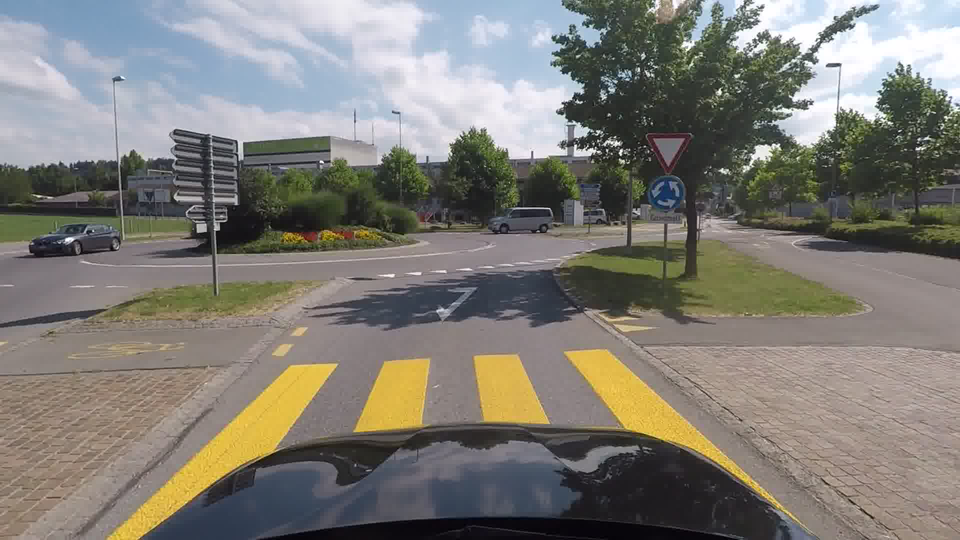}
        \includegraphics[width=0.495\linewidth]{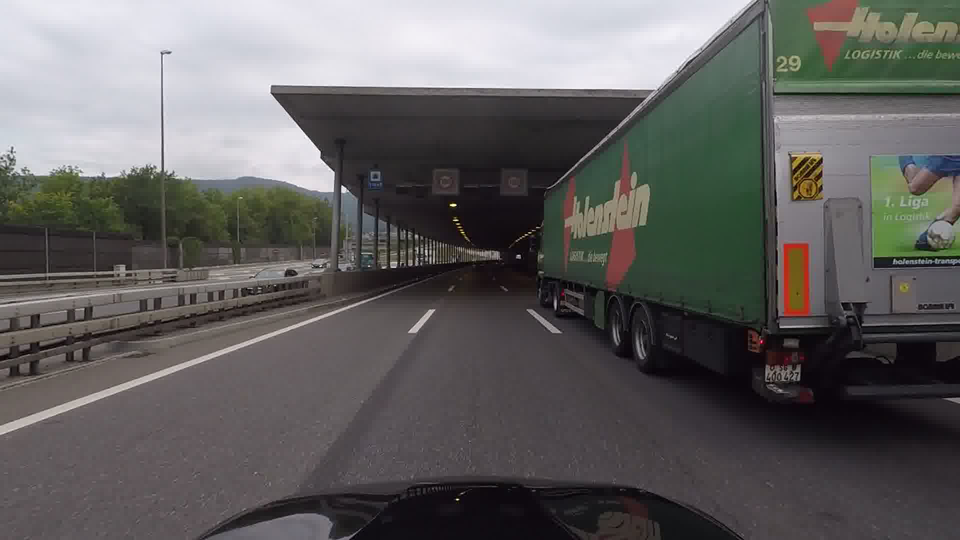}
        \includegraphics[width=0.495\linewidth]{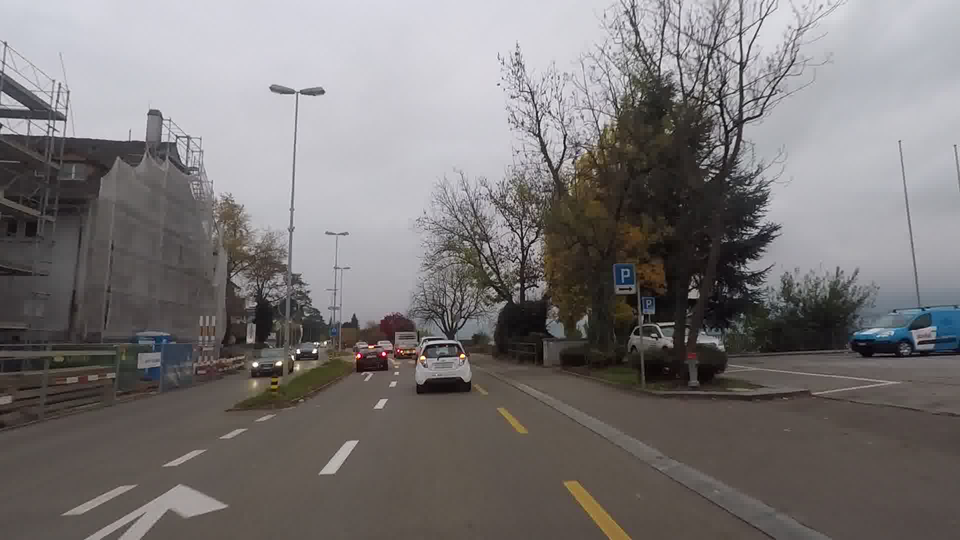}
    \caption{Sample of training images.}
    \label{fig:train}
\end{figure}

\begin{figure}
    \centering
        \includegraphics[width=0.495\linewidth]{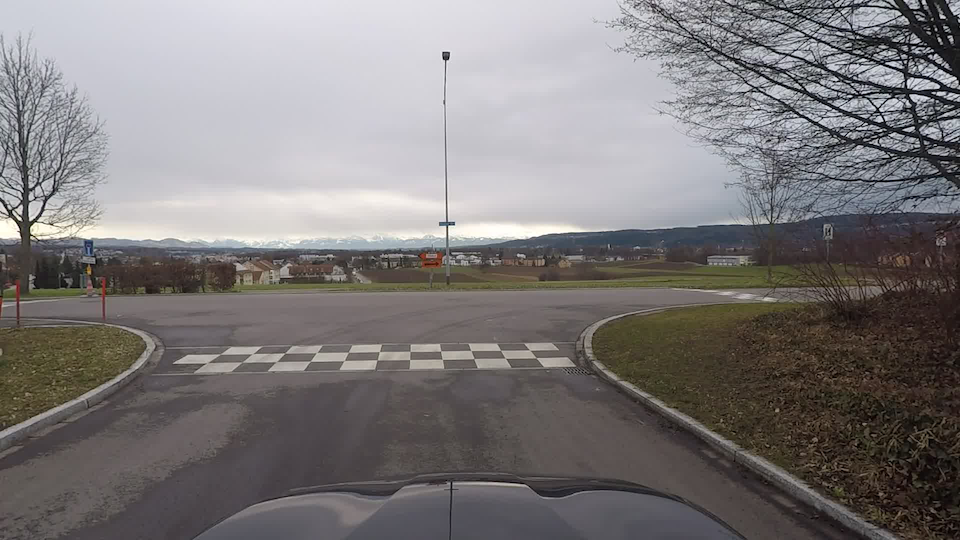}
        \includegraphics[width=0.495\linewidth]{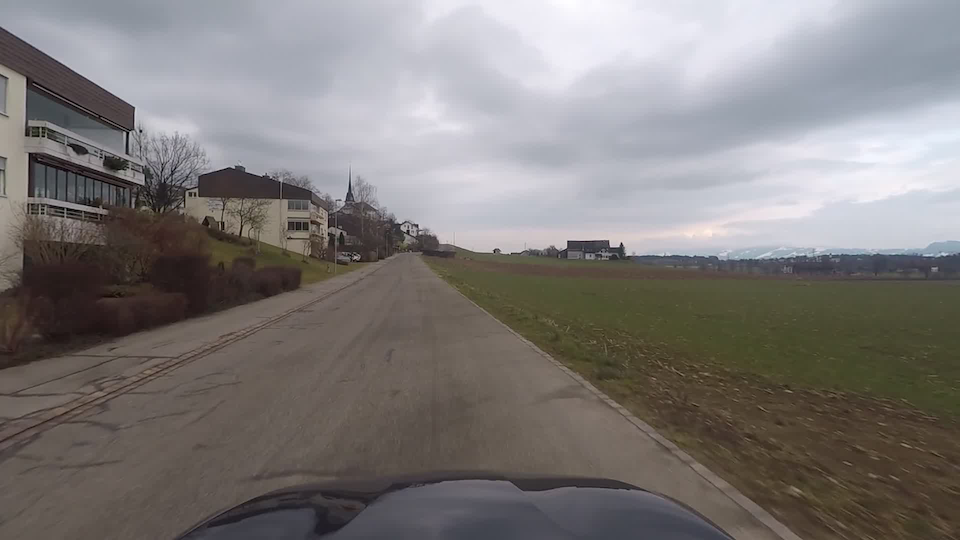}
        \includegraphics[width=0.495\linewidth]{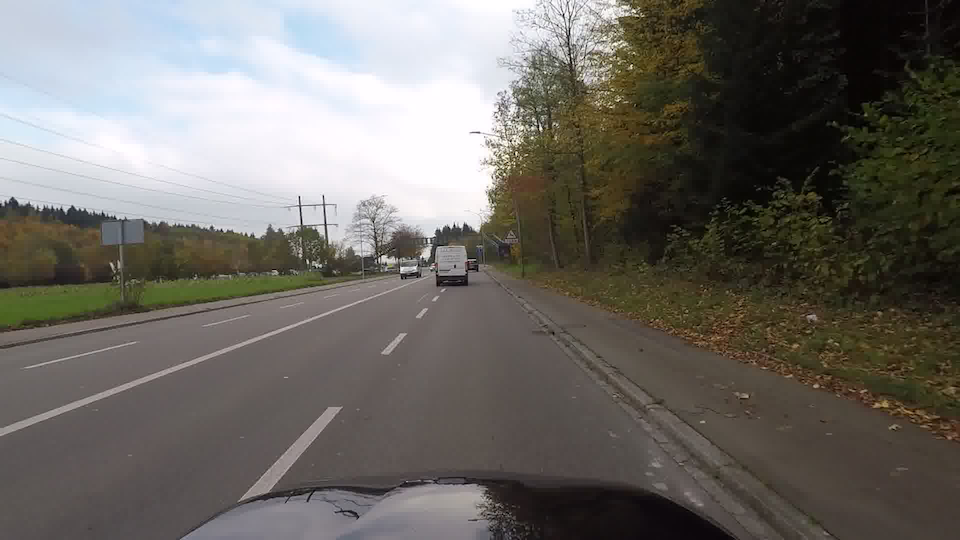}
        \includegraphics[width=0.495\linewidth]{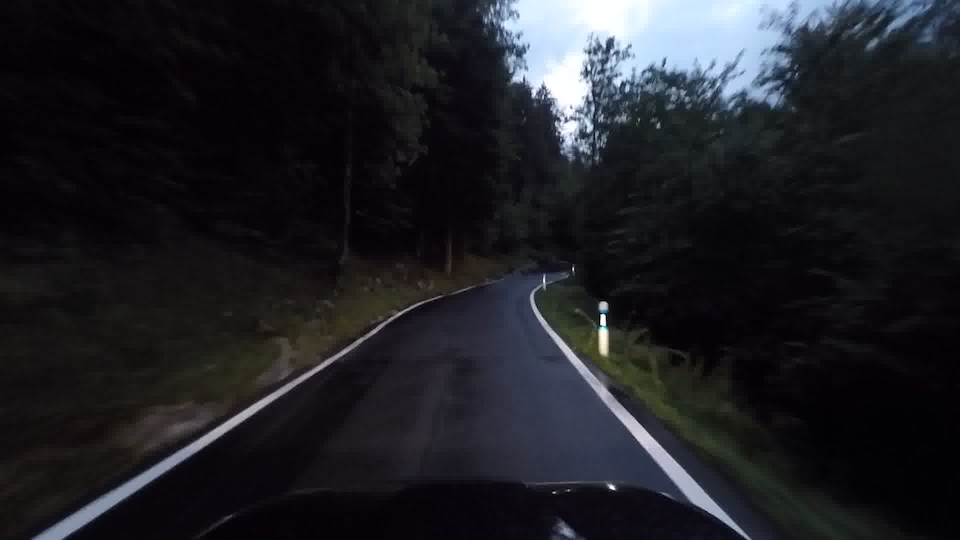}
        \includegraphics[width=0.495\linewidth]{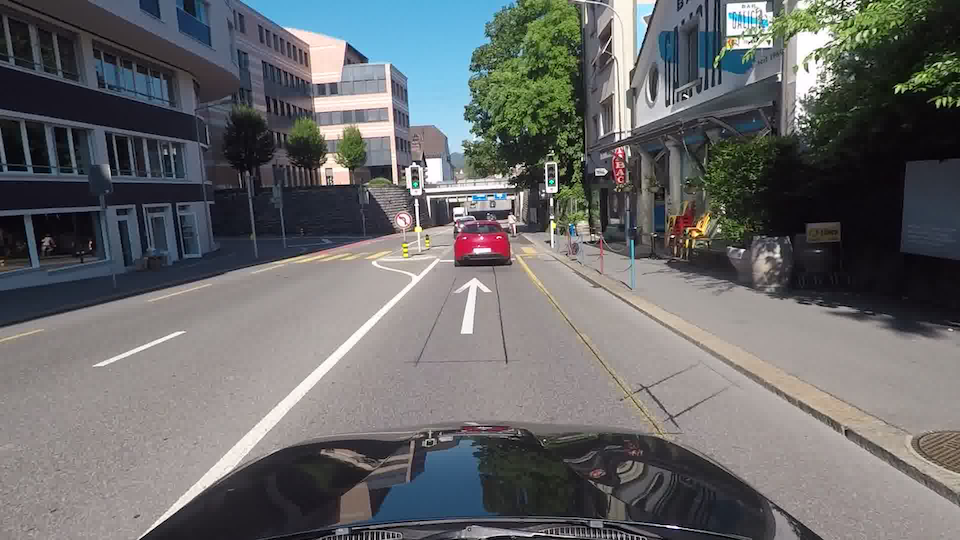}
        \includegraphics[width=0.495\linewidth]{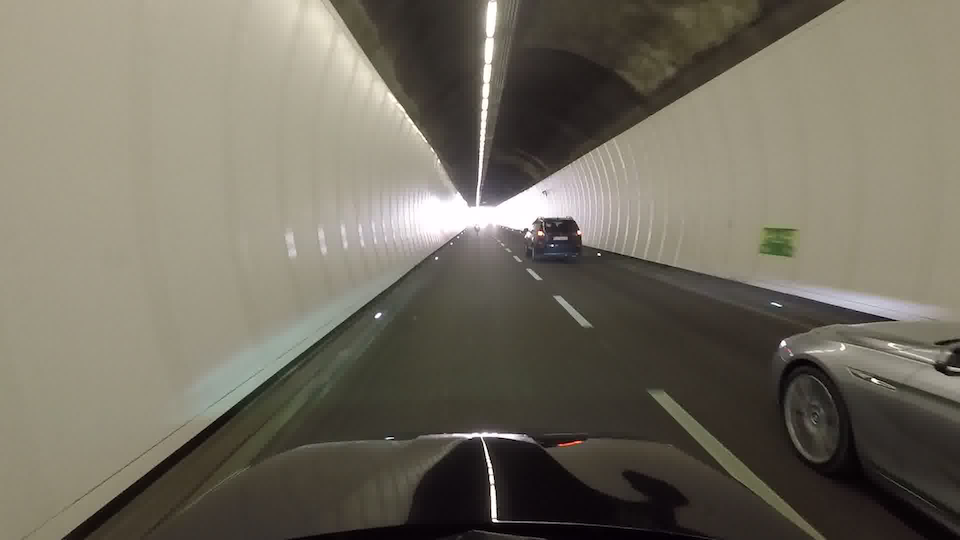}
        \includegraphics[width=0.495\linewidth]{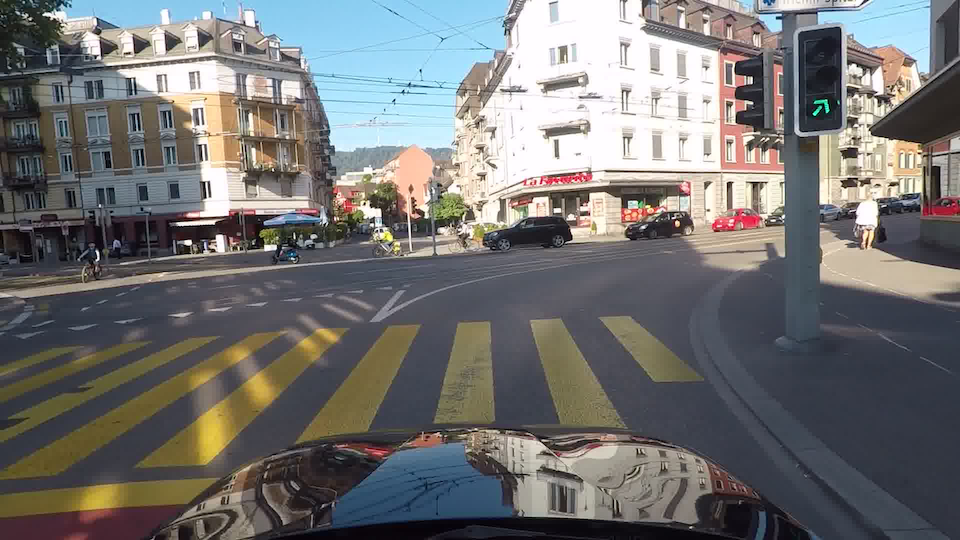}
        \includegraphics[width=0.495\linewidth]{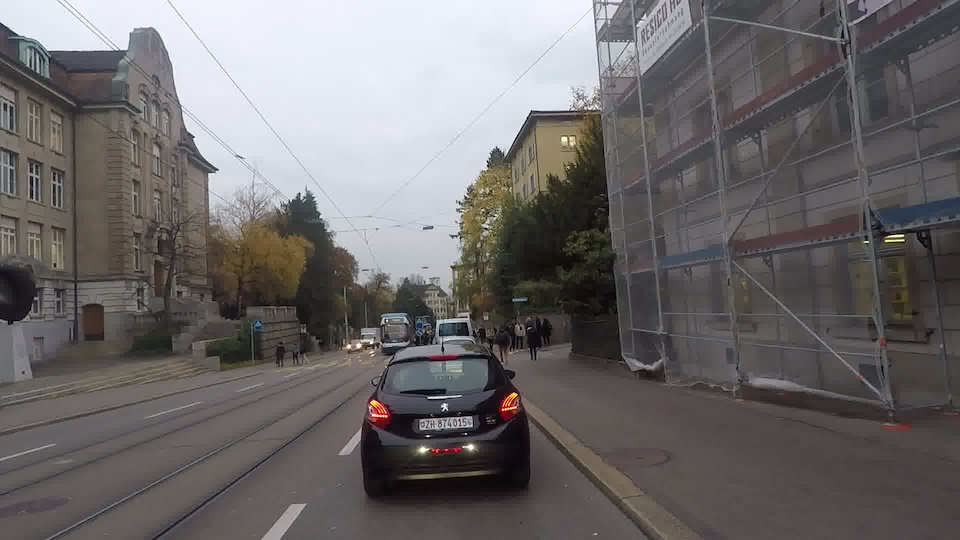}
        \includegraphics[width=0.495\linewidth]{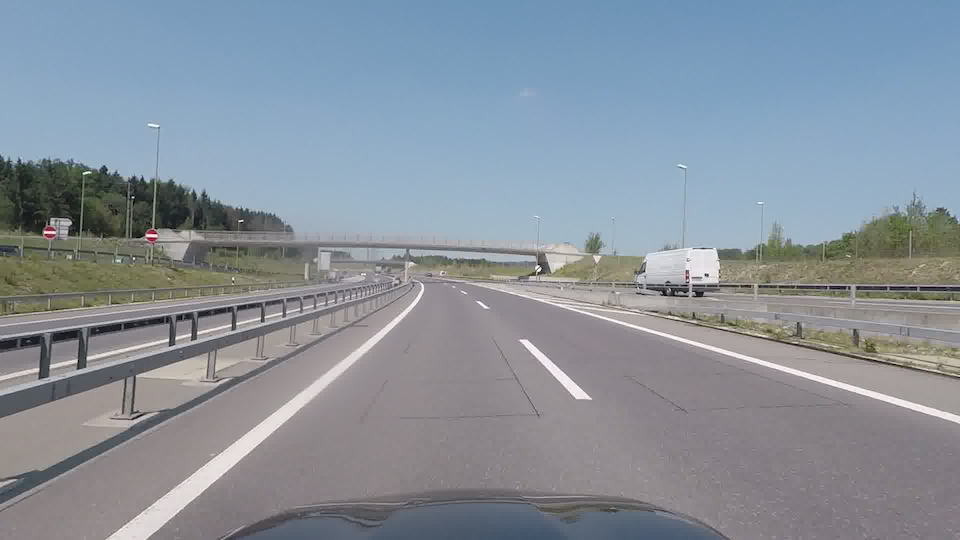}
        \includegraphics[width=0.495\linewidth]{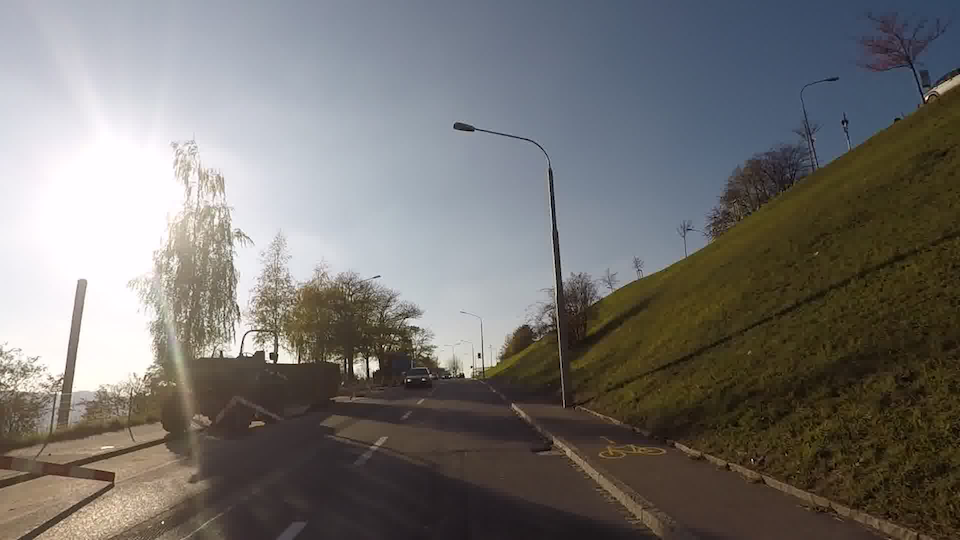}
    \caption{Sample of test images.}
    \label{fig:test}
\end{figure}

The challenge dataset consists of videos of around 55 hours of recorded driving in Switzerland, and the associated driving speed and steering angle. Sample training images are shown in Figure \ref{fig:train} and sample testing images are shown in Figure \ref{fig:test}. The dataset consists of about 2 million images taken using a GoPro Hero5 facing the front, sides, and rear of a car driven around Switzerland. An image of a visual map from HERE technologies and a semantic map derived from this data is provided. The semantic map consists of 21 fields. GPS latitude, GPS longitude, and GPS precision. All images and data are sampled at 10 frames per-second. The data is separated into 5 minute chapters. In total, there are 682 chapters for 27 different routes. The data is randomly sampled into 548 chapters for training, 36 chapters for validation, and 98 chapters for testing. 

\section{Methods}
\label{sec:methods}

We pre-process the data by down-sampling, normalization, semantic map imputation and augmentation as described next:

\paragraph{Down-sampling}
For efficient experimentation we down-sample the dataset in both space and time as shown in Table \ref{tab:sampling}. Although the initial process is time consuming, it enabled the model to train faster by limiting I/O time. The initial image size is also prohibitive due to memory limitations on our GPUs and significantly decreases the speed at which we train the model. 

\begin{table}
\small
\centering
\begin{tabular}{l|llll}
Dataset & Full & Sample 1 & Sample 2 & Sample 3 \\
\hline
Res.  & 1920x1080 & 640x360  & 320x180  & 160x90   \\
Train & 1,600,000 & 160,000     & 80,000      & 40,000\\
Val.  & 106,000   & 10,600    & 5,300     & 2,650    \\
Test  & 290,000      & 29,000      & 14,500    & 7,250 \\
\hline
\end{tabular}
\caption{Data sampling in both space and time for efficient experimentation.}
\label{tab:sampling}
\end{table}

\paragraph{Network architecture}

\begin{figure}
    \includegraphics[width=1\linewidth]{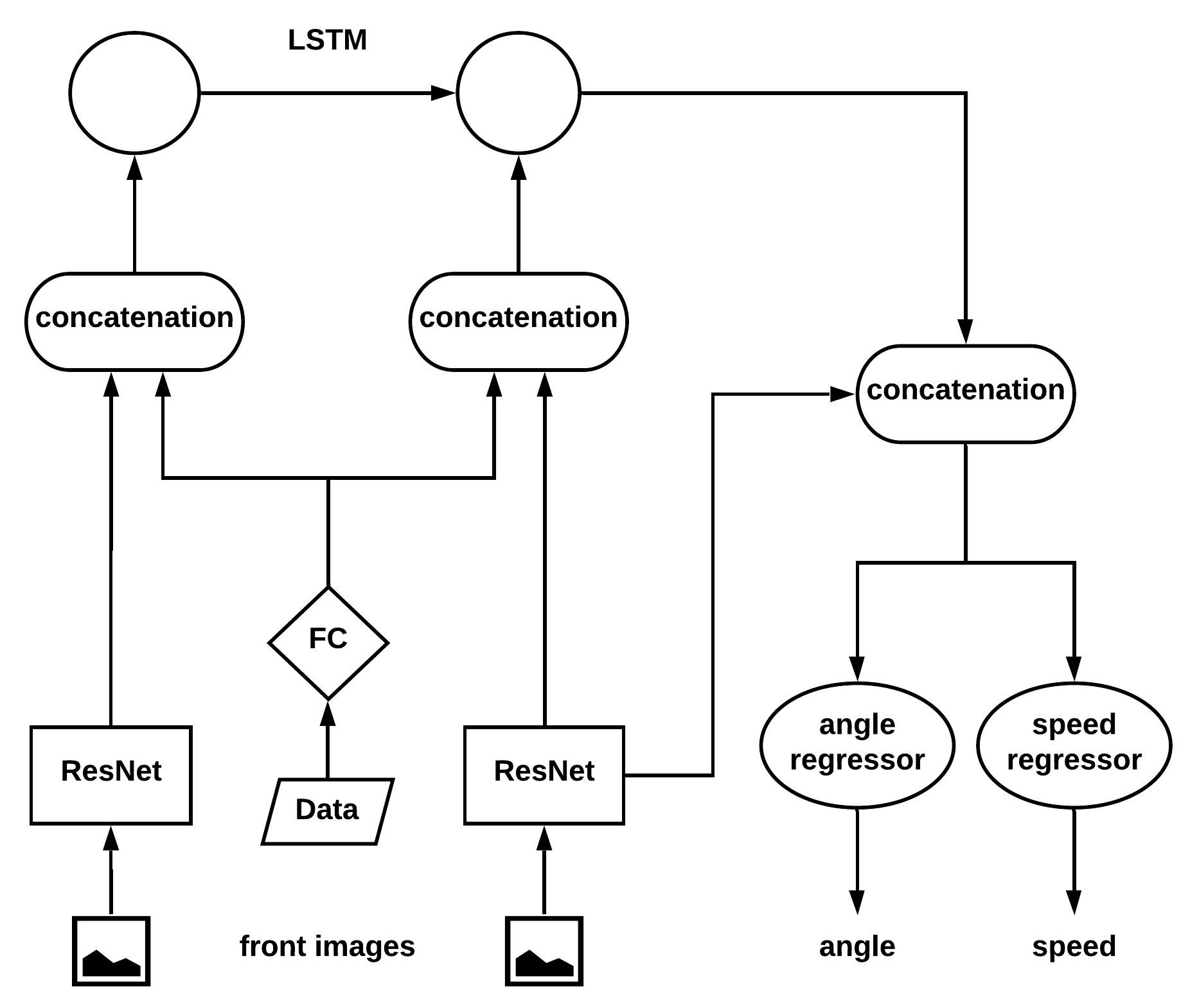}
    \caption{Deep neural network architecture: The network consists of a pre-trained ResNet and fully connected network that feeds into an LSTM model. This and the output of the ResNet model on the current image are fed into an angle and speed regressor.}
    \label{fig:architecture}
\end{figure}

\begin{figure}
    \includegraphics[width=1\linewidth]{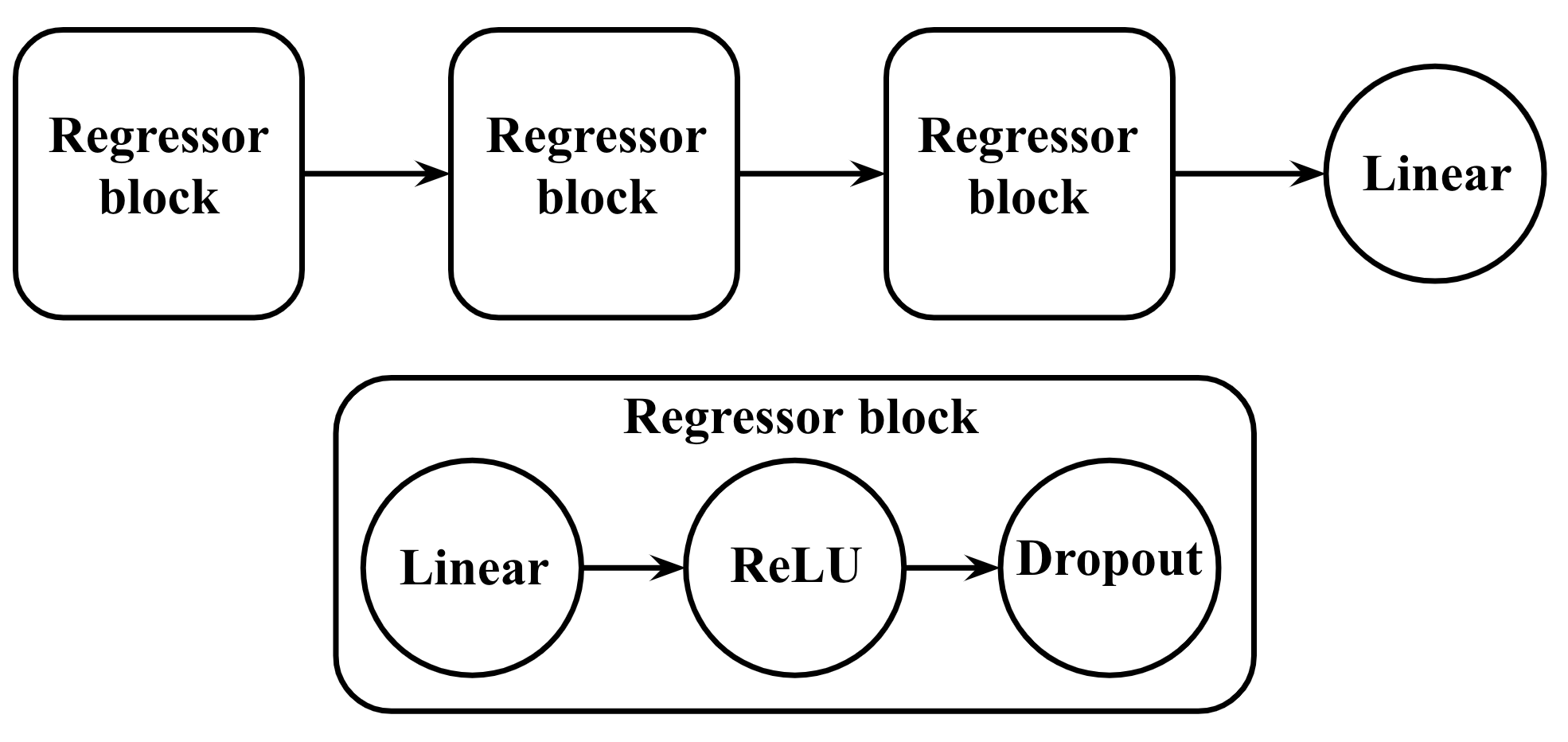}
    \caption{Speed and steering angle regressors: Each regressor consists of 3 blocks of a linear layer, a ReLu activation, and a 10\% droput. The hidden layers are 1024, 512, and 256.}
    \label{fig:regressors}
\end{figure}

Figure \ref{fig:architecture} shows our network architecture. The inputs consist of the front facing GoPro \cite{gopro} images and the semantic map from HERE Technologies \cite{here}. We include the current and previous frame in each iteration, which are 0.4 seconds apart. The images are fed into a pre-trained ResNet34 model or a ResNet152 model. We feed the semantic map into a fully connected network with two hidden layers of 256, 128 and with ReLU activation layers. The output from the fully connected and ResNet models are concatenated and fed into a long short-term memory (LSTM) network. The output from the LSTM network and the current information from the semantic map, if used, are concatenated. The LSTM output and the current information is then used as input to both an angle regressor and a speed regressor, which predict the current steering angle and speed. Both regressors share the same structure, as shown in Figure \ref{fig:regressors}.

\paragraph{Normalization.} The images are normalized in all models according to the default vectors of mean and standard deviation provided. In a similar fashion, the steering angle and speed are normalized for training using means and standard deviations. 

\paragraph{Semantic map.} We use 20 of the numerical features as shown in Table \ref{tab:features}. Missing values are imputed in with zeros. We avoid using the mean value across the chapter or other placeholder to avoid using future data. For one model, we add in 27 additional dummy variables for each folder an image was located. Although the semantic map has information on location, the information from the folder gave another view that we consider possibly helpful to the training process. 

\begin{table}
\small
\centering
\begin{tabular}{l|l}
hereMmLatitude     & hereSegmentExitHeading  \\
hereMmLongitude    & hereSegmentEntryHeading \\
hereSpeedLimit     & hereCurvature           \\
hereSpeedLimit\_2  & hereCurrentHeading      \\
hereFreeFlowSpeed  & here1mHeading           \\
hereSignal         & here5mHeading           \\
hereYield          & here10mHeading          \\
herePedestrian     & here20mHeading          \\
hereIntersection   & here50mHeading          \\
hereMmIntersection & hereTurnNumber         
\end{tabular}
\caption{Features used from the Semantic map from HERE Technologies.}
\label{tab:features}
\end{table}

\paragraph{Ensemble.} We digitize the steering angles and speeds from the training dataset into 100 and 30 evenly spaced bins, where each bin represents a range of continuous values for angle or speed. Then we count the digitized values for each bin and use these counts to estimate the likelihood of an angle or speed value. Using this estimated distribution, we compute a weighted average of predictions from a selection of models we trained. Table \ref{tab:results} shows the performance of each individual model and the ensemble. We combine predictions from epoch 1 of model 3 - 5 for angle and predictions from epoch 1 of model 2 - 5 and epoch 2 of model 1 for speed. 


\subsection{Implementation}

\paragraph{Hyper-parameters of network training.}
All models are trained using the Adam optimizer with an initial learning rate of 0.0001, without weight decay, and momentum parameters $\beta_1 = 0.9$, and $\beta_2 = 0.999$. In all models, we optimize on the sum of mean squared error (MSE) for the steering angle and speed. Attempts at optimizing only one regressor showed little improvement to either metrics. Minibatch sizes are either 8, 32, or 64, limited by GPU memory. The image set, depending on run, is of dimensions of 320x180 or 160x90 and using ResNet34 or ResNet152 models for the image CNN. Table \ref{tab:results} summarizes the hyperparameters and image settings used for each of our models.

\paragraph{Computation} 
Preprocessing took about 3 days of continuous computation on an Intel i7-4790k CPU using 6 of the 8 cores and with data stored on a 7200rpm hard drive. The largest limitation to the preprocessing was the I/O due to the sizeable number of photos. Models 1, 2, 4, and 5 were trained using a single Nvidia K80 GPU. Each epoch required approximately 12 hours to complete. Model 3 was trained on a single Nvidia GTX 980 GPU and each epoch took approximately 10 hours to compute. The code is implemented in Python 3.6 using PyTorch.

\section{Results}
\label{sec:results}

\begin{table*}
\small
\centering
\begin{tabular}{lccccc|ccc}
Model    & CNN       & Image Dimensions & Semantic Map & Batch & Epochs & MSE Angle & MSE Speed & Combined \\
\hline
1        & ResNet34  & 320x180          & No           & 64    & 2      & 1111.437 & 5.866 & 1117.303 \\
\hline
2        & ResNet152 & 320x180          & No           & 8     & 1      & 1211.434 & 5.461 & 1216.895 \\
\hline
3        & ResNet34  & 160x90           & 20 Features  & 8     & 1      & 897.489  & 6.664 & 904.153  \\
         &           &                  &              &       & \textbf{2}      & \textbf{883.501}  & \textbf{6.403} & \textbf{889.904}  \\
         &           &                  &              &       & 3      & 931.689  & 6.445 & 938.134  \\
         &           &                  &              &       & 4      & 970.96   & 6.714 & 977.674  \\
         &           &                  &              &       & 5      & 956.262  & 6.576 & 962.838  \\
\hline
4        & ResNet34  & 320x180          & 20 Features  & 32    & 1      & 995.42   & 5.316 & 1000.736 \\
         &           &                  &              &       & 2      & 946.516  & 5.337 & 951.853  \\
         &           &                  &              &       & 3      & 989.013  & 5.519 & 994.532  \\
         &           &                  &              &       & 4      & 965.791  & 5.706 & 971.497  \\
         &           &                  &              &       & 5      & 987.572  & 5.846 & 993.418  \\
\hline
5        & ResNet34  & 320x180          & 47 Features  & 64    & 1      & 900.407  & 5.571 & 905.978  \\
\hline
Ensemble &           &                  &              &       &        & \textbf{831.504}  & \textbf{4.543} & \textbf{836.047} \\
\hline
\end{tabular}
\caption{Parameters and results for 5 different models and the result of the ensemble. The best overall result is an ensemble of the single models. Individually, we note that the inclusion of the semantic map reduces the MSE by about 300 (comparing models 1 and 4) and using the smaller image size resulted in an additional benefit (comparing models 3 and 4). Models 3 and 4 likely suffer from overfitting as evidence by the increasing test MSE, although the training loss decreased, as shown in Figure \ref{fig:trainloss}. The best standalone and overall models are in bold.}
\label{tab:results}
\end{table*}

\begin{figure}
    \includegraphics[width=\linewidth]{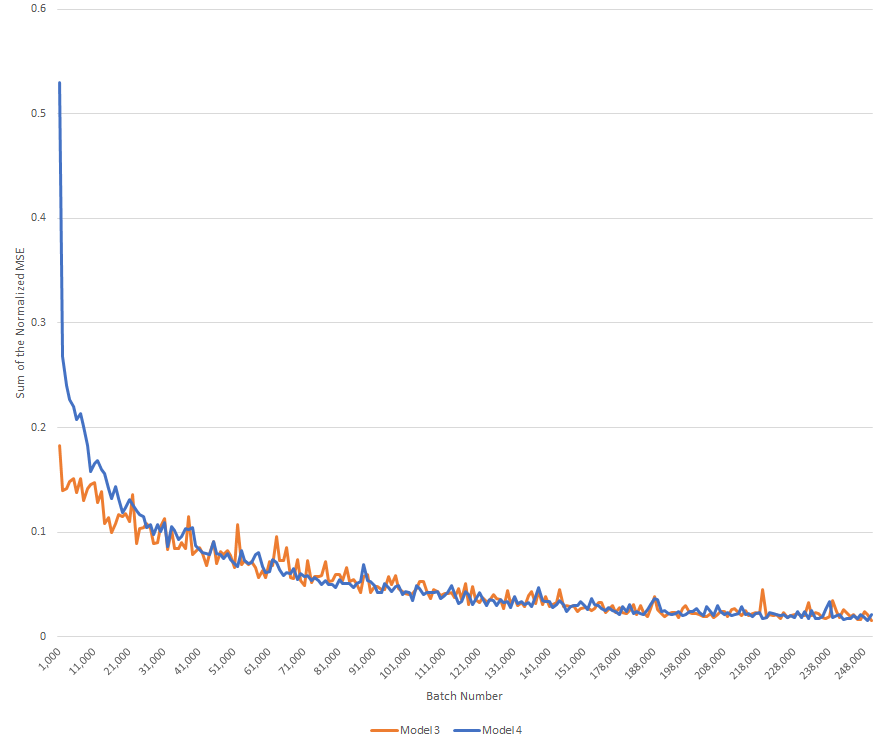}
    \caption{Training loss of angle and speed MSE as a function of epoch.}
    \label{fig:trainloss}
\end{figure}

\begin{table}
\small
\centering
\begin{tabular}{l|cc}
\hline
Model         & MSE Angle & MSE Speed \\
\hline
Overall       & 831.5    & 4.5   \\
Zone30        & 2,981.1  & 0.3   \\
Zone50        & 1,353.4  & 6.0   \\
Zone80        & 168.6    & 4.1   \\
Right         & 1,928.4  & 1.3   \\
Straight      & 821.7    & 4.6   \\
Left          & 833.6    & 2.6   \\
Pedestrian    & 3,722.1  & 4.1   \\
Traffic Light & 329.2    & 5.3   \\
Yield         & 1,818.9  & 2.8  \\
\hline
\end{tabular}
\caption{Performance across various zones: the ensemble method did worst in the pedestrian sections and best in Zone80. Presumably, pedestrian sections would be hardest to train due to the unpredictability of cities and people. Zone80 sections are likely straighter and require less change in speed and steering angle and would probably be easier to train. Likewise, Right and Left sections, which would require learning a turn, would be difficult, but Straight segments are easier to train and did better. }
\label{tab:mse}
\end{table}
We train each of our five neural network models for varying number of epochs to make efficient use of our computational resources. Table \ref{tab:results} summarizes our results. The most significant improvement on the single models occurs by including of the HERE semantic map into the model, which results in a decrease of the MSE by approximately 300 points on the Angle metric. This is most apparent between models 1 and 4. Including the city location as part of the semantic map has a marginal benefit to the model, as seen by the change from model 4 to 5. Notably, models 3 and 4 tend to overfit. Our best result for both is after 2 epochs, and the MSE slowly increases for each additional epoch. The training loss for both models decrease throughout the training, as shown in Figure \ref{fig:trainloss}. The performance of the ensemble method in the different road types is shown in Table \ref{tab:mse}.

\section{Conclusions}
\label{sec:conclusions}
In conclusion, fusing the semantic map data with the image data significantly improves results. We demonstrate that fusing these different modalities can be efficiently performed using a simple neural network. A classical ensemble method of diverse models improves overall performance. We make our models and code publicly available \cite{diodato2019learningtodrive}.

\paragraph{Acknowledgements} We would like to thank the Columbia University students of the Fall 2019 Deep Learning class for their participation in the challenge. Specifically, we would like to thank Xiren Zhou, Fei Zheng, Xiaoxi Zhao, Yiyang Zeng, Albert Song, Kevin Wong, and Jiali Sun for their participation.

\newpage
\clearpage

{\small
\bibliographystyle{ieee_fullname}
\bibliography{bibliography}
}

\end{document}